\documentclass{article}
\usepackage{spconf,amsmath,graphicx,pdflscape,amsmath}
\usepackage{times}
\usepackage{epsfig}
\usepackage{amssymb}
\usepackage{multirow}
\usepackage{enumitem}
\usepackage[accsupp]{axessibility}
\usepackage{spconf,amsmath,graphicx,pdflscape,amsmath}
\usepackage{ragged2e}
\usepackage{url} 

\begin{document}

\title{GRU-AUNet: A Domain Adaptation Framework for Contactless Fingerprint Presentation Attack Detection}
%
\name{Banafsheh Adami, Nima Karimian\thanks{This material is in part based on work supported by the National Science Foundation CAREER Award under Grant No. 2338981.}}
\address{West Virginia University}

\maketitle

\begin{abstract}

Although contactless fingerprints offer user comfort, they are more vulnerable to spoofing. The current solution for anti-spoofing in the area of contactless fingerprints relies on domain adaptation learning, limiting their generalization and scalability. To address these limitations, we introduce GRU-AUNet, a domain adaptation approach that integrates a Swin Transformer-based UNet architecture with GRU-enhanced attention mechanisms, a Dynamic Filter Network in the bottleneck, and a combined Focal and Contrastive Loss function. Trained in both genuine and spoof fingerprint images, GRU-AUNet demonstrates robust resilience against presentation attacks, achieving an average BPCER of 0.09\% and APCER of 1.2\% in the CLARKSON, COLFISPOOF, and IIITD datasets, outperforming state-of-the-art domain adaptation methods.
\end{abstract}


\section{Introduction}
\label{sec:intro}
Biometric systems have found extensive utility across various domains, including but not limited to law enforcement and forensics, singular identification, healthcare, and facilitating access control for smartphones and tablets. These applications contribute to enhanced convenience in our day-to-day activities. The demand for contactless biometric solutions is increasing rapidly due to hygiene-related issues. Fingerprints and facial biometrics are recognized as the primary modalities in the field of biometrics which extensive implementation by law enforcement agencies and national ID programs on a global scale~\cite{jain201650}. According to the biometric system market is projected to reach a value of \$82.9 billion by 2027~\cite{bworld}. Despite popularity of face authentication, it has encountered challenges during the pandemic, particularly regarding the use of face coverings~\cite{calbi2021consequences}, which prevents its high rate of 
detection~\cite{sun2013deep,yang2022mask,botezatu2022fun}.
Contactless fingerprint recognition provides great potential in various applications, offering a touchless and hygienic biometric authentication solution. Contactless fingerprinting is a cutting-edge technological advancement in the field of biometrics that eliminates the need for traditional bioscanner sensors~\cite{grosz2021c2cl,lin2018cnn,lin2018matching,kumar2013towards,cui2023monocular}. Instead, it relies solely on a smartphone camera lens for capturing and recording fingertip information. Compared to touch based fingerprint, it is considered to be more seamless and convenient and has a higher user acceptance.

\begin{figure}
    \centering
    \includegraphics[width=0.95\linewidth]{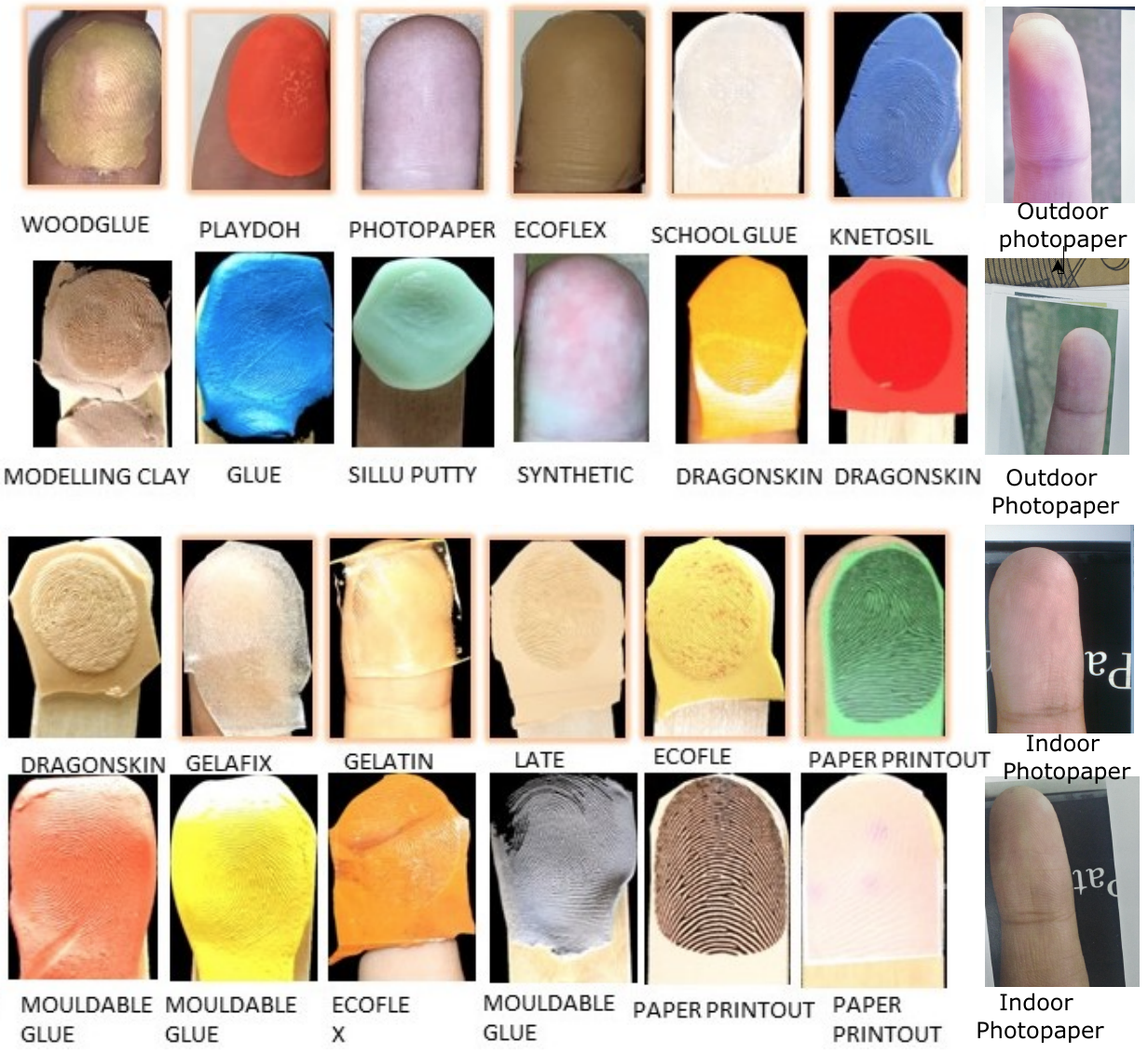}    
    \caption{Different spoofed samples from CLARKSON ,COLFISPOOD, and III-TD datasets were employed in this paper. First four spoof samples are related to CLARKSON dataset (WOODGLUE,PLAYDOH, PHOTOPAPER, ECOFLEX), and others are spoof samples from COLFISPOOF dataset, indoor and outdoor photopaper are realted to III-TD dataset.}
    \label{fig:PAI}
\end{figure}
Although contactless fingerprint technology provides convenient and widely accepted user experiences, it does come with various drawbacks. These include lower biometric performance, susceptibility to environmental influences, and vulnerabilities to presentation attacks~\cite{kolberg2023colfispoof}. Presentation attacks can compromise the security and reliability of biometric authentication systems, potentially leading to unauthorized access or identity theft~\cite{tolosana2019biometric}. Hence, developing an effective countermeasure against contactless fingerprint is crucial to detect and prevent any unseen presentation attacks. Contactless fingerprint systems used smartphones for capturing photo-based finger images are more vulnerable to spoofing due to the use of a single type of camera and limited computational capabilities (Figure.~\ref{fig:PAI} shows fingerprint spoof images fabricated for this study). While well-studied has been conducted on contact-based fingerprint recognition and its vulnerabilities to presentation attacks, limited attention has been given to studying contactless fingerprint presentation attacks. In recent years, only few numbers of approaches have been explored to identify various contactless fingerprint presentation attacks using hand-crafted features, and deeply learned features.
Deep learning models have been widely applied in biometrics and other domains to improve feature extraction and classification~\cite{hosseini2024intra}. However, despite significant progress in contactless fingerprint PAD, several limitations remain, including:

\begin{itemize}
  \item Disparities in data distributions: existing PAD approaches on contactless fingerprint assume similar data distributions between training and testing scenarios. However, this assumption leads to limited generalization capabilities of PAD methods when faced with real-world situations, especially with unseen attacks.
  \item Multiple types of presentation attacks: contactless fingerprint presentation attacks can take various forms, including printed attack, latex, ecoflex, and more. Thus, generating and creating a labeled training set that covers all possible presentation attacks for each new application scenario is impractical.
\end{itemize}
Our work introduces a novel unsupervised learning approach based on the Swin-UNet architecture with attention mechanisms for contactless fingerprint anti-spoofing. The proposed method demonstrates improved generalization, scalability, and robustness against diverse presentation attacks, as validated through comprehensive experiments on multiple datasets. Our research makes a significant contribution to the field of biometric security by enhancing the security and reliability of contactless fingerprint recognition systems.
The main contributions of our work are:
\begin{itemize}
    \item Development of a novel unsupervised learning approach for contactless fingerprint anti-spoofing: We have proposed a new method that combines the Swin-UNet architecture with a Swin Transformer backbone to detect presentation attacks in contactless fingerprint recognition systems. This approach is unsupervised, meaning it is trained solely on genuine fingerprint images without the need for labeled spoof samples during training.
    \item Improved generalization and scalability: By training the model only on genuine fingerprint images, our approach addresses the limitations of existing methods that rely on mixed training data of genuine and spoof samples. This unsupervised learning strategy enhances the model's ability to generalize to unseen presentation attacks and scales better to new application scenarios.
    \item Comprehensive evaluation: The proposed method has been evaluated on multiple datasets which contains a wide range of presentation attacks, demonstrating its robustness and generalization ability. The comparison with state-of-the-art approaches and the reported performance metrics (APCER, BPCER, and HTER) suggest that the proposed method is effective.
    \item Robustness against various presentation attacks: Our model has been evaluated against a diverse range of presentation attacks, including those from the CLARKSON, COLFISPOOF, and IIITD Spoofed Fingerphoto databases. These databases cover a wide variety of spoofing techniques, such as printed attacks, ecoflex, playdoh, gelatin, and more. Our model's ability to detect these various attacks demonstrates its robustness and effectiveness.
    \item Integration of attention mechanisms and skip connections: We have modified the Swin-UNet architecture by replacing the skip connections with a novel attention path. This attention path incorporates channel attention and spatial attention to help the model focus on the most relevant features for accurate classification. The integration of attention mechanisms and skip connections enhances the model's ability to capture discriminative features and improves its overall performance.
    \item Comprehensive evaluation and comparison: We have conducted extensive experiments to evaluate the performance of Our proposed method and compared it with state-of-the-art approaches. The results demonstrate the superiority of our Swin-UNet-based approach in terms of APCER, BPCER, and HTER metrics across multiple datasets. This comprehensive evaluation validates the effectiveness of our unsupervised learning approach for contactless fingerprint anti-spoofing.
    \item Contribution to the field of biometric security: Our work addresses a critical challenge in contactless fingerprint recognition systems, which are increasingly popular due to their convenience and hygienic advantages. By developing an effective anti-spoofing solution that can detect various presentation attacks, our research contributes to enhancing the security and reliability of contactless fingerprint authentication systems, thereby advancing the field of biometric security.
\end{itemize}

\section{Related Work}
We have summarized the previous works on contactless fingerprint technology in the Table~\ref{tab:previous works}.
Fujio et al.\cite{fujio2018face} were early pioneers in investigating the use of deep neural networks for contactless fingerprint anti-spoofing, achieving an impressive half-error rate of 0.04\%. Marasco et al.\cite{marasco2021fingerphoto} utilized Convolutional Neural Network (CNN) architectures, including ResNet and AlexNet, on the IIITD Spoofed Finger Photo Database, achieving a Detection Equal Error Rate (D-EER) of 2.14\% for AlexNet and 0.96\% for ResNet. They later made slight improvements compared to their baseline approach~\cite{marasco2021deep}. However, it's worth noting that the ResNet model was trained on both live and spoofed images, potentially limiting its representativeness of real-world scenarios and scalability. In 2022, they introduced a method to bolster the PAD system's resilience against color paper print-out attacks~\cite{marasco2022late}, employing a U-Net with a ResNet-50 backbone for photo segmentation. This effort resulted in an impressive APCER of 0.1\% with a BPCER of 2.67\%.

Kolberg et al.\cite{kolberg2023colfispoof} introduced the COLFISPOOF dataset tailored for non-contact fingerprint Presentation Attack Detection (PAD) tasks, consisting of 7200 samples covering 72 distinct types of spoofed attacks captured using two different smartphone devices. Purnapatra et al.\cite{purnapatra2023presentation} utilized DenseNet-121 and NasNetMobile models with a newly available public database, incorporating both live and spoof data in their training and achieving an APCER of 0.14\% and a BPCER of 0.18\%. Hailin Li et al.~\cite{li2023deep} demonstrated the effectiveness of presentation attack detection (PAD) using various models, including AlexNet, DenseNet-201, MobileNet-V2, NASNet, ResNet50, and Vision Transformer. Among these, the Vision Transformer achieved the best APCER and BPCER. Their study encompassed over 5,886 genuine samples and 4,247 spoof samples, considering four distinct training cases for different types of spoof for testing. Notably, the ResNet50 model achieved an 8.6\% equal error rate (EER). However, despite the promising performance during training, these recent models exhibited limited generalization to new counterfeit images, resulting in suboptimal performance in such scenarios.
Puranpatra et al.~\cite{purnapatra2023liveness} organized a competition on fingerprint liveness detection, where the winning solution achieved various APCER values such as 9.20\% for paper printout, 0\% for ecoflex, playdoh, and latex, 0.1\% for wood glue, and 99.9\% for synthetic fingertip at BPCER=0.62\%.
\begin{table*}[htb]
\begin{center}
\resizebox{\textwidth}{!}{%
\begin{tabular}{|l|l|l|l|l|l|l|}

\hline
Author & Year & Method & Database & Spoof type  & Results \\ \hline
Tanej et al.~\cite{taneja2016fingerphoto} & 2016 & Hand crafted & \begin{tabular}[c]{@{}l@{}}IIITD: \\ class: 128 \\ images: 5100 \end{tabular} & \begin{tabular}[c]{@{}l@{}}Print Attack\\ Photo Attack\end{tabular}   & EER = 3.71\% \\ \hline
Wasnik et al.~\cite{wasnik2018presentation} & 2018 &\begin{tabular}[c]{@{}l@{}} Hand crafted LBP,\\ BSIF, HOG, SVM \end{tabular}& \begin{tabular}[c]{@{}l@{}} subjects: 50 \\images: 250 \\  videos: 150\end{tabular}  & \begin{tabular}[c]{@{}l@{}}print artefact\\ electronic replay\\ elctronic display\end{tabular}  & \begin{tabular}[c]{@{}l@{}}BPCER = 1.8, 0, 0.66,\\ APCER = 10\end{tabular} \\ \hline
Fujito et al.~\cite{fujio2018face}& 2018 & AlexNet & \begin{tabular}[c]{@{}l@{}}Live: 4096\\ spoofe sample: 8192\end{tabular} &  \begin{tabular}[c]{@{}l@{}}Print Attack \\ Photo Attack\end{tabular}  & HTER = 0.04\% \\ \hline
Marasco et al.~\cite{marasco2021fingerphoto}, \cite{marasco2021deep} & 2022 & \begin{tabular}[c]{@{}l@{}}AlexNet DenseNet201,\\  ResNet18,DenseNet121,\\ ResNet34, MobileNEt-V2\end{tabular} & IIITD  &\begin{tabular}[c]{@{}l@{}} Print Attack\\Photo Attack\end{tabular} &  \begin{tabular}[c]{@{}l@{}}D-EER\_AlexNet = 2.14\\ D-EER\_ResNet = 0.96\%\end{tabular} \\ \hline

Kolberg et al.~\cite{kolberg2023colfispoof}& 2023 & Not Reported & \begin{tabular}[c]{@{}l@{}}COLFISPOOF:\\ 7200 spoof samples \\ 72 different PAI\end{tabular} & \begin{tabular}[c]{@{}l@{}}Knetosil, Mould glue, \\ latex, silly putty,\\ paper printout, \\ school glue, \\ dragonskin, \\ ecoflex, gelatin, \\ glue,  modelling clau, \\ playdoh\end{tabular}  & not reported \\ \hline
Purnapatra et al.~\cite{purnapatra2023presentation} & 2023 & \begin{tabular}[c]{@{}l@{}}DenseNet 121,\\ NASNet\end{tabular} & \begin{tabular}[c]{@{}l@{}}35 subjects with 12 devices\\ attack sample: 7548 \\ synthetic: 10000\end{tabular}  & \begin{tabular}[c]{@{}l@{}}ecoflex, playdoh,\\ wood glue, \\ synthetic, fingerphoto,\\ latex\end{tabular}  &\begin{tabular}[c]{@{}l@{}} APCER = 0.14\% \\ BPCER = 0.18\% \end{tabular}\\ 
\hline

Hailin Li et al.\cite{li2023deep} & 2023 & \begin{tabular}[c]{@{}l@{}}AlexNet,DenseNet201,\\MobileNet-V2,ResNet50 \\ NasNet, GoogleNet,\\EfficientNEt-B0\\Vision Transformer\end{tabular} & \begin{tabular}[c]{@{}l@{}}5886 bonafide\\ 4247 attack sample\\ four PAIs types\end{tabular}  & \begin{tabular}[c]{@{}l@{}}ecoflex, playdoh,\\ wood glue, \\ synthetic, fingerphoto, \\latex\end{tabular}  &\begin{tabular}[c]{@{}l@{}} They report APCER and BPCER\\in 4 cases,\\in each case one PAI\\used only for testing \\and three remains \\used for training \end{tabular}\\ \hline

Puranpatra et.al.\cite{purnapatra2023liveness} & 2023 & \begin{tabular}[c]{@{}l@{}}Combination of \\two CNN\end{tabular} & \begin{tabular}[c]{@{}l@{}}5886 bonafide\\ 4247 attack sample\\ four PAI types \end{tabular}  & \begin{tabular}[c]{@{}l@{}}ecoflex, playdoh,\\ wood glue, \\  fingerphoto, \\latex\end{tabular}  &\begin{tabular}[c]{@{}l@{}} BPCER = 0.62\\ APCER = 11.35\\ ACER = 6 \end{tabular}\\ \hline

B Adami et al.\cite{adami2023universal} & 2023 & \begin{tabular}[c]{@{}l@{}}Resnet-18/LeakyRelu,\\(Combined Loss)\end{tabular} & \begin{tabular}[c]{@{}l@{}}5886 bonafide\\ 4247 attack sample\\ 10,000 synthetic \end{tabular}  & \begin{tabular}[c]{@{}l@{}}ecoflex, playdoh,\\ wood glue, \\ synthetic, fingerphoto, \\latex\end{tabular}  &\begin{tabular}[c]{@{}l@{}} BPCER = 0.12\\ APCER = 0.63\\ ACER = 0.68 \end{tabular}\\ \hline

B Adami et al.\cite{adami2024contactless}  & 2024 & \begin{tabular}[c]{@{}l@{}}convolution autoencoder,\\ CBAM-autoencoder,\\Swin-transformer\end{tabular} & \begin{tabular}[c]{@{}l@{}}35 subjects with 12 devices\\ attack sample: 7548 \\ synthetic: 10000\end{tabular}  & \begin{tabular}[c]{@{}l@{}}ecoflex, playdoh, wood glue, \\ synthetic, fingerphoto, latex\end{tabular}  &\begin{tabular}[c]{@{}l@{}} APCER = 1.6\% \\ BPCER = 0.96\% \end{tabular}\\ \hline
\end{tabular}%
}
\caption{Summary of previous works for contactless fingerprint anti-spoofing. HOG-- histogram of oriented gradients (HOG), SVM-- support vector machine, LBP--local binary patterns, BSIF--binarized statistical image features, EER -- equal error rate, TAR -- true acceptance rate, FAR -- false acceptance rate BPCER--bonafide presentation classification error rate, HTER -- half total error rate, APCER-- attack presentation classification error rate.}
\label{tab:previous works}
\end{center}
\end{table*}

\section{Proposed Method}
\begin{figure*}[htb]
\centering
\includegraphics[width=0.9\linewidth]{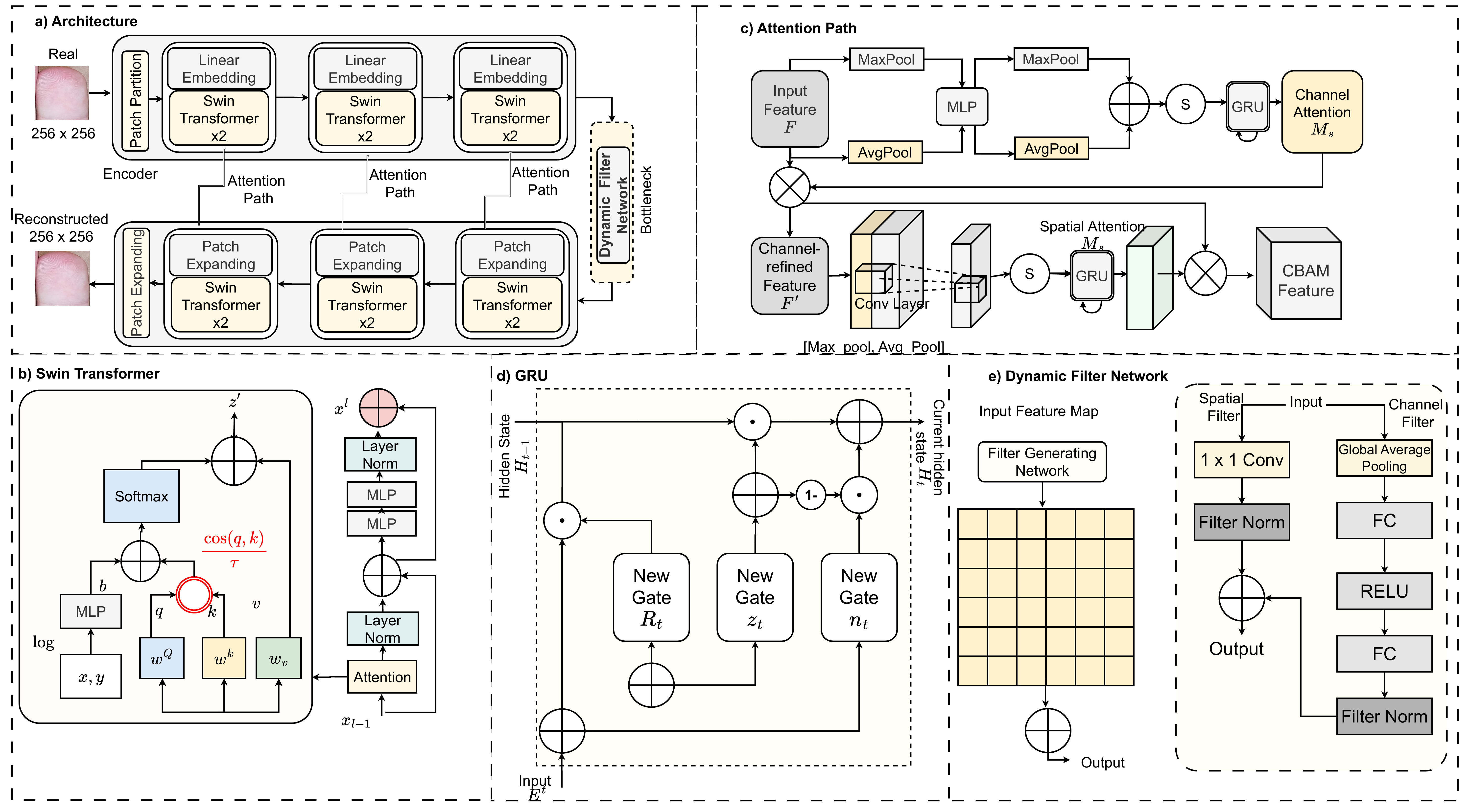}
\caption{\textbf{GRU-AUNet Architecture:} (a) The encoder processes $256 \times 256$ RGB inputs, dividing them into patches and tokenizing into $C$-dimensional vectors. (b) These vectors pass through Swin Transformer blocks, patch merging layers, and the Dynamic Filter Network in the bottleneck, which dynamically adapts filter responses. (c) The decoder mirrors this structure, upsampling features to recover spatial resolution for anti-spoofing detection. (d) A attention path replaces skip connections, focusing on critical features to improve classification accuracy. (e) In the bottleneck, the Dynamic Filter Network processes input feature maps through spatial and channel filter branches. The spatial branch applies a $1 \times 1$ convolution, while the channel branch uses global average pooling, fully connected layers, and ReLU activations.}
\label{fig:gru_attendunet_architecture}
\end{figure*}

The proposed method introduces GRU-AUNet, an advanced architecture designed for contactless fingerprint anti-spoofing. This architecture integrates a Swin Transformer V2~\cite{liu2021swin} backbone with an encoder-decoder structure enhanced by GRU-based attention mechanisms and a novel Dynamic Filter Network (DFN) in the bottleneck, as illustrated in Figure~\ref{fig:gru_attendunet_architecture}. This combination allows the model to effectively learn and differentiate between genuine and spoof fingerprint features, improving its generalization and robustness against a wide range of presentation attacks (PAs).

\subsection{Swin Transformer}
The Swin Transformer V2 serves as the backbone for feature extraction, providing hierarchical processing of input fingerprint images~\cite{liu2022swin}. The model divides each image into nonoverlapping patches and embeds these patches into high-dimensional feature vectors. These vectors undergo successive layers of self-attention and patch merging operations, which capture complex spatial dependencies and hierarchical features within the data. The self-attention mechanism is based on a scaled cosine similarity function:

\begin{equation}
\begin{gathered}
\text{Attention} = \operatorname{softmax}\left(\frac{QK^T}{\sqrt{d_z}} + B_{ij}\right) V \\
\operatorname{Sim}(q_i, k_j) = \frac{\text{cos}(q_i, k_j)}{\tau} + B_{ij}
\end{gathered}
\end{equation}

where \(QK^T\) represents the projection between the query vectors (\(Q\)) and key (\(K\)) vectors, and \(\sqrt{d_z}\) is a scaling factor for normalization. The term \(B_{ij}\) accounts for the relative position bias between the pixels \(i\) and \(j\), and \(\tau\) is a scalar that can be learned that modulates the attention distribution.

\subsection{Dynamic Filter Network in Bottleneck}
The core innovation of our architecture is the incorporation of a Dynamic Filter Network (DFN) in the bottleneck. The DFN dynamically adjusts its filtering operations based on the input feature characteristics, enabling the model to capture intricate patterns that are critical for distinguishing between genuine and spoofed fingerprints~\cite{jia2016dynamic}. The DFN operates through two parallel branches: the spatial filter branch and the channel filter branch.

The spatial filter branch applies a \(1 \times 1\) convolution followed by filter normalization, focusing on local spatial features. The channel filter branch, on the other hand, employs Global Average Pooling (GAP), followed by fully connected (FC) layers and ReLU activations, to generate channel-specific filters. The outputs from both branches are then combined to produce a refined feature map, which is further normalized. The dynamic filter operation can be formulated as:

\begin{equation}
F_{dynamic}(x) = \sum_{i=1}^{n} \alpha_i(x) * W_i
\end{equation}

where \(x\) is the input feature map, \(W_i\) represents the learnable filter weights, and \(\alpha_i(x)\) denotes the dynamically generated filter coefficients conditioned on the input \(x\). This adaptability enhances the model's ability to tailor its feature extraction process to the specific characteristics of each input, as shown in Figure~\ref{fig:gru_attendunet_architecture}-e.

\subsection{Attention Path with GRU-Enhanced Mechanisms}
The attention path in GRU-AUNet replaces the traditional skip connections found in standard UNet architectures. This novel path integrates both channel-wise and spatial attention mechanisms, allowing the model to dynamically adjust its focus to the most critical features throughout the network~\cite{dey2017gate}. GRU-based attention mechanisms further refine this process by leveraging recurrent connections to maintain temporal coherence across different levels of the network, leading to improved classification accuracy, particularly in distinguishing between real and spoof fingerprints.

\subsection{Attention Classifier}
The final classification layer of GRU-AUNet is designed to efficiently distinguish between genuine and spoof fingerprints. It comprises convolutional layers followed by attention blocks, each enhanced with GRU-based attention refinement. Attention blocks use Convolutional Block Attention Modules (CBAM)~\cite{woo2018cbam} with modifications to incorporate GRU functions. The GRU refinement process can be expressed as:

\begin{equation} \label{eq.GAN}
\begin{aligned}
    z_t = \sigma(W_z \cdot [h_{t-1}, x_t]) \\ 
    r_t = \sigma(W_r \cdot [h_{t-1}, x_t]) \\
    n_t = \tanh(W_n \cdot [r_t \odot h_{t-1}, x_t]) \\
    h_t = (1 - z_t) \odot n_t + z_t \odot h_{t-1}
\end{aligned}
\end{equation}

where \(z_t\) is the update gate, \(r_t\) is the reset gate, \(n_t\) is the new gate, and \(h_t\) is the updated hidden state. The learnable weight matrices \(W_z\), \(W_r\), and \(W_n\) control the GRU's operation, while the sigmoid (\(\sigma\)) and hyperbolic tangent (\(\tanh\)) activation functions ensure non-linearity and stability in the attention refinement process.

\begin{figure}[htb]
\centering
\includegraphics[width=1.0\linewidth]{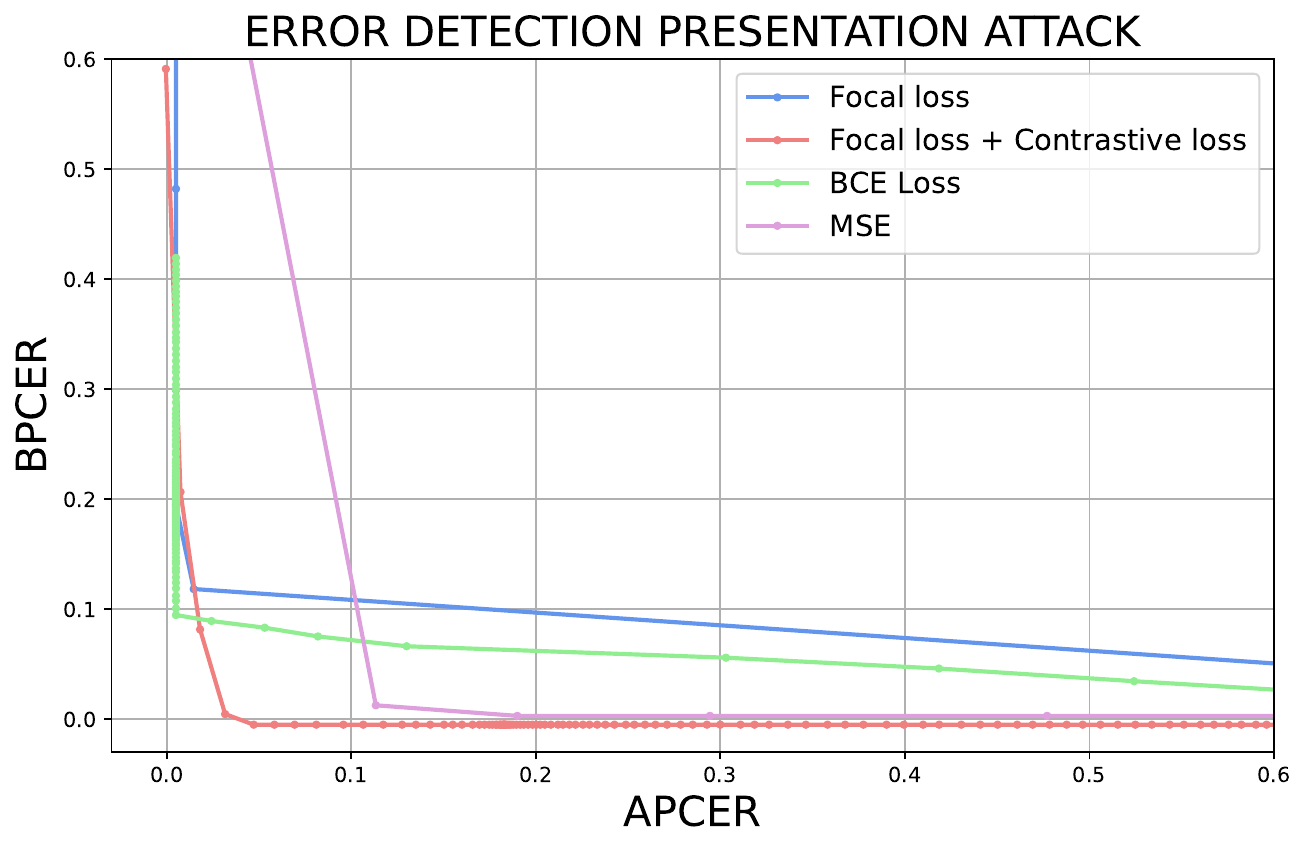}
\caption{Comparison of spoofing detection model performance using different loss functions, with APCER (attack samples misclassified as genuine) on the x-axis and BPCER (genuine samples misclassified as attacks) on the y-axis.}
\label{fig:roc}
\end{figure}

\begin{table*}[htb]
\centering
\huge 
\resizebox{\linewidth}{!}{%
\begin{tabular}{|l|l|c|c|c|l|}
\hline
\textbf{Dataset}  & \textbf{Method} & \textbf{APCER (\%)} & \textbf{BPCER (\%)} & \textbf{ACER (\%)} & \textbf{Spoof Types (APCER \%)} \\ \hline
\multirow{3}{*}{CLK} 
                  & DenseNet-121~\cite{purnapatra2023presentation} & 88.03 & 0.18 & 44.11 & Eco (0), PH (88.03), PL (0.14), WO (0) \\
                  & NASNetMobile~\cite{purnapatra2023presentation}  & 82.15 & 9.04 & 45.60 & Eco (0), PH (82.15), PL (0.71), WO (5.96)D \\
                  & ResNet~\cite{adami2024contactless}&
                  88.03 & 0.18 & 44.11 & Eco (0), PH (9.43), PL (0.14), WO (0) \\
                  & \textbf{GRU-AUNet (Ours)}  & \textbf{1.2}  & \textbf{0.09} & \textbf{0.65} & Eco (0), PH (0), PL (17.3), WO (0) \\ \hline
                  
\multirow{3}{*}{CFS} DF
                  & ResNet-50~\cite{li2023deep}  & 45.33 & 3.33 & 24.33 & GEL (13.32), GFX (NA), GLU (NA), LAT (NA), MC (NA) \\
                  & DenseNet-201~\cite{li2023deep} & 94.72 & 3.33 & 49.03 & GEL (15.33), GFX (NA), GLU (NA), LAT (NA), MC (NA) \\
                  &ResNet~\cite{adami2024contactless} & 0 & .09 & 0 & GEL (0), GFX (0), GLU (0), LAT (0), MC (0) \\
                  & \textbf{GRU-AUNet (Ours)}  & \textbf{0.0} & \textbf{0.09} & \textbf{0.04} & GEL (0), GFX (0), GLU (0), LAT (0), MC (0) \\ \hline
\multirow{3}{*}{IIITD} 
                  & ResNet18~\cite{marasco2021fingerphoto} & NA & NA & NA & PA (NA), PH (NA), PL (NA) \\
                  & DenseNet-121~\cite{marasco2021deep} & NA & NA & NA & PA (NA), PH (NA), PL (NA) \\
                  & \textbf{GRU-AUNet (Ours)}  & \textbf{0.21 (PH)} & \textbf{0.09} & \textbf{0.15} & 0.0 (PA), PA (0), PH (0.21), PL (0) \\ \hline
\end{tabular}%
}
\caption{Performance of GRU-AUNet compared to state-of-the-art methods across different datasets, with APCER values for each type of spoof material. Dataset Abbreviations: CLK = CLARKSON, CFS = COLFISPOOF, IIITD = IIITD Spoofed Fingerphoto Database. Spoof Type Abbreviations: Eco = Ecoflex, PH = Photopaper, PL = Playdoh, WO = Woodglue, GEL = Gelatin, GFX = Gelafix, GLU = Glue, LAT = Latex, MC = Mouldable Clay, PA = Printed Attack.}
\label{tab:individual_results}
\end{table*}

\section{Experimental Setup}
\subsection{Loss Function and Optimization}
To train the model, we utilize a labeled dataset comprising both genuine and spoof fingerprint images. The model's parameters are optimized using the Adam optimizer~\cite{kingma2014adam} in conjunction with a learning rate scheduler.

We combine the Focal Loss~\cite{lin2017focal} and Contrastive Loss~\cite{hadsell2006dimensionality} to enhance the model's ability to learn discriminative features for differentiating between real and spoof fingerprints. The combined Focal Contrastive Loss is defined as: $\mathcal{L}_{FC} = \mathcal{L}_{\text{Focal}} + \lambda \mathcal{L}_{\text{Contrastive}}$, where $\mathcal{L}_{\text{Focal}}$ represents the Focal Loss, $\mathcal{L}_{\text{Contrastive}}$ represents the Contrastive Loss, and $\lambda$ is a hyperparameter that balances the contributions of the two components of loss. The Focal Loss addresses class imbalance and emphasizes difficult examples~\cite{lin2017focal}:

\begin{align}
\mathcal{L}_{\text{Focal}}(y, \hat{y}) = & -\frac{1}{N} \sum_{i=1}^{N} \left[ \alpha (1 - \hat{y}_i)^{\gamma} \cdot y_i \cdot \log(\hat{y}_i) \right. \nonumber \\
& \left. + (1 - y_i) \cdot (1 - \alpha \hat{y}_i)^{\gamma} \cdot \log(1 - \hat{y}_i) \right] 
\end{align}

Here, $y_i$ is the ground truth binary label (0 for real, 1 for spoof) for the $i$-th sample, $\hat{y}_i$ is the predicted probability of the $i$-th sample being spoofed, and $N$ is the total number of samples. Parameters $\alpha$ and $\gamma$ are hyperparameters that adjust the importance of positive versus negative examples and focus on hard-to-classify samples, respectively.

The Contrastive Loss aims to promote the learning of distinguishable features~\cite{hadsell2006dimensionality}:

\begin{align}
\mathcal{L}_{\text{Contrastive}} = \frac{1}{2N} \sum_{i=1}^{N} \left[ y_i \frac{1}{2} d_i^2 + (1 - y_i) \frac{1}{2} \max(0, m - d_i)^2 \right]
\end{align}
In this equation, $y_i$ is the binary label indicating whether the $i$-th pair is similar ($y_i = 1$) or dissimilar ($y_i = 0$). The term $d_i$ represents the Euclidean distance between the feature embeddings of the $i$-th pair, while $m$ is the margin hyperparameter that determines the minimum distance between dissimilar pairs. The Contrastive Loss encourages the model to learn an embedding space where genuine fingerprints are clustered together and spoofed fingerprints are distinctly separated.

Combining Focal Loss and Contrastive Loss enhances the model's ability to handle contactless fingerprint anti-spoofing. Focal Loss addresses class imbalance by focusing on difficult examples, while Contrastive Loss improves feature discrimination between genuine and spoof fingerprints. Together, they boost classification accuracy and robustness in detecting presentation attacks.

\subsection{Database}
We used three publicly available datasets: CLARKSON~\cite{purnapatra2023presentation}, COLFISPOOF~\cite{kolberg2023colfispoof}, and IIITD Spoofed Fingerphoto Database ~\cite{sankaran2015smartphone} 
,~\cite{taneja2016fingerphoto}. These datasets include a wide range of spoofing techniques and materials, providing a comprehensive evaluation of our model. Each dataset contains labeled samples of both genuine and spoof fingerprints, facilitating domain adaptation training and evaluation.

\subsection{Evaluation Metrics}
The model's performance was evaluated using Bonafide Presentation Classification Error Rate (BPCER), Attack Presentation Classification Error Rate (APCER), and Average Classification Error Rate (ACER). These metrics, along with the Receiver Operating Characteristic (ROC) curve, were used to assess the model's ability to distinguish between live and spoof data.

\subsection{Results}
We evaluated the performance of the trained GRU-AUNet model on the CLARKSON, COLFISPOOF, and IIITD Spoofed Fingerphoto Database datasets, which include both live and spoof fingerprint samples. The model was trained using labeled data from these datasets, including both genuine and various spoof samples, to effectively learn the distinguishing features necessary for accurate classification.

Table~\ref{tab:individual_results} presents the performance metrics of our GRU-AUNet model on these datasets, including comparisons with several state-of-the-art methods such as DenseNet-121~\cite{purnapatra2023presentation}, NASNetMobile~\cite{purnapatra2023presentation}, ResNet-50~\cite{li2023deep}, DenseNet-201~\cite{li2023deep}, and EfficientNet-B0~\cite{li2023deep}. Our model achieves an impressive average BPCER of 0.09\% and an APCER of 1.2\% on the CLARKSON dataset, outperforming existing domain adaptation learning methods. The model also demonstrates robust performance across various presentation attack instruments (PAIs), including ECOFLEX, PHOTOPAPER, PLAYDOH, and WOODGLUE.

\subsection{Cross-Dataset Validation}
To further evaluate the generalization capability of our model, we conducted cross-dataset validation. The GRU-AUNet model trained on the CLARKSON dataset was tested on the COLFISPOOF and IIITD datasets, achieving an APCER of 0.4\% and a BPCER of 0.11\% on the COLFISPOOF dataset. Similarly, when trained on the IIITD dataset and tested on CLARKSON, the model achieved an APCER of 0.3\% and a BPCER of 0.12\%, demonstrating the model's robust generalization across diverse datasets.

\begin{table}[htb]
\centering
\resizebox{\linewidth}{!}{%
\begin{tabular}{|l|c|c|c|}
\hline
\textbf{Training Dataset} & \textbf{Testing Dataset} & \textbf{APCER (\%)} & \textbf{BPCER (\%)} \\ \hline
CLARKSON & COLFISPOOF  & 0.4 & 0.11 \\ \hline
IIITD    & CLARKSON    & 0.3 & 0.12 \\ \hline
COLFISPOOF & CLARKSON  & 0.5 & 0.10 \\ \hline
\end{tabular}%
}
\caption{Cross-dataset validation results for GRU-AUNet.}
\label{tab:cross_dataset}
\end{table}

\subsection{K-Fold Cross-Validation}
To further ensure the reliability and stability of our model's performance, we applied 5-fold cross-validation on each dataset. The GRU-AUNet model consistently achieved high performance across all folds, with an average APCER of 1.3\% and BPCER of 0.08\%. This validation technique underscores the robustness of the model, indicating that it maintains high accuracy even when trained and tested on different subsets of the same dataset.

\begin{table}[htb]
\centering
\resizebox{\linewidth}{!}{%
\begin{tabular}{|l|c|c|c|c|c|c|}
\hline
\textbf{Dataset} & \textbf{Fold 1} & \textbf{Fold 2} & \textbf{Fold 3} & \textbf{Fold 4} & \textbf{Fold 5} & \textbf{Average} \\ \hline
CLARKSON~\cite{purnapatra2023presentation} & 1.5\% & 1.2\% & 1.1\% & 1.3\% & 1.4\% & 1.3\% \\ \hline
COLFISPOOF~\cite{kolberg2023colfispoof} & 0.09\% & 0.07\% & 0.08\% & 0.09\% & 0.08\% & 0.08\% \\ \hline
IIITD~\cite{sankaran2015smartphone,taneja2016fingerphoto}& 0.21\% & 0.19\% & 0.20\% & 0.21\% & 0.22\% & 0.21\% \\ \hline
\end{tabular}%
}
\caption{K-Fold Cross-Validation results for GRU-AUNet across different datasets.}
\label{tab:kfold_results}
\end{table}

\section{Conclusion}
\label{sec:conclude}
In this work, we proposed GRU-AUNet, an advanced domain adaptation framework for contactless fingerprint presentation attack detection. Our model integrates a Swin Transformer-based UNet architecture with GRU-enhanced attention mechanisms and a Dynamic Filter Network (DFN) in the bottleneck. This architecture, combined with a Focal and Contrastive Loss function, effectively improves the model’s generalization and robustness against presentation attacks across multiple datasets. Comprehensive evaluations on the CLARKSON, COLFISPOOF, and IIITD Spoofed Fingerphoto databases demonstrated that GRU-AUNet significantly outperforms existing state-of-the-art methods. Our model achieved an APCER of 1.2\% and a BPCER of 0.09\% on the CLARKSON dataset, improving upon prior methods such as DenseNet-121 and NASNetMobile, which reported APCERs of 88.03\% and 82.15\%, respectively. Similar superior performance was observed on COLFISPOOF and IIITD datasets, where our model maintained high accuracy in detecting a diverse range of presentation attack instruments (PAIs). Additionally, cross-dataset validation results confirmed the model’s strong generalization capabilities, with an APCER as low as 0.3\% when trained on IIITD and tested on CLARKSON.

\section{Acknowledgments}
This project was supported in part by the National Science Foundation under Grants No. 2338981.

\bibliographystyle{IEEEbib}
{\small
\bibliography{strings,refs}

\begin{thebibliography}{10}

\bibitem{jain201650}
Anil~K Jain, Karthik Nandakumar, and Arun Ross,
\newblock ``50 years of biometric research: Accomplishments, challenges, and opportunities,''
\newblock {\em Pattern recognition letters}, vol. 79, pp. 80--105, 2016.

\bibitem{bworld}
MultiMedia LLC,
\newblock ``{Biometric Market Size},'' \url{https://www.fortunebusinessinsights.com/biometric-system-market-107100}, 2023,
\newblock [Online; accessed 22-June-2023].

\bibitem{calbi2021consequences}
Marta Calbi, Nunzio Langiulli, Francesca Ferroni, Martina Montalti, Anna Kolesnikov, Vittorio Gallese, and Maria~Alessandra Umilta,
\newblock ``The consequences of covid-19 on social interactions: an online study on face covering,''
\newblock {\em Scientific reports}, vol. 11, no. 1, pp. 2601, 2021.

\bibitem{sun2013deep}
Yi~Sun, Xiaogang Wang, and Xiaoou Tang,
\newblock ``Deep convolutional network cascade for facial point detection,''
\newblock in {\em Proceedings of the IEEE conference on computer vision and pattern recognition}, 2013, pp. 3476--3483.

\bibitem{yang2022mask}
Chun-Wei Yang, Thanh~Hai Phung, Hong-Han Shuai, and Wen-Huang Cheng,
\newblock ``Mask or non-mask? robust face mask detector via triplet-consistency representation learning,''
\newblock {\em ACM Transactions on Multimedia Computing, Communications, and Applications (TOMM)}, vol. 18, no. 1s, pp. 1--20, 2022.

\bibitem{botezatu2022fun}
Cristian Botezatu, Mathias Ibsen, Christian Rathgeb, and Christoph Busch,
\newblock ``Fun selfie filters in face recognition: Impact assessment and removal,''
\newblock {\em IEEE Transactions on Biometrics, Behavior, and Identity Science}, vol. 5, no. 1, pp. 91--104, 2022.

\bibitem{grosz2021c2cl}
Steven~A Grosz, Joshua~J Engelsma, Eryun Liu, and Anil~K Jain,
\newblock ``{C2CL} : Contact to contactless fingerprint matching,''
\newblock {\em IEEE Transactions on Information Forensics and Security}, vol. 17, pp. 196--210, 2021.

\bibitem{lin2018cnn}
Chenhao Lin and Ajay Kumar,
\newblock ``A cnn-based framework for comparison of contactless to contact-based fingerprints,''
\newblock {\em IEEE Transactions on Information Forensics and Security}, vol. 14, no. 3, pp. 662--676, 2018.

\bibitem{lin2018matching}
Chenhao Lin and Ajay Kumar,
\newblock ``Matching contactless and contact-based conventional fingerprint images for biometrics identification,''
\newblock {\em IEEE Transactions on Image Processing}, vol. 27, no. 4, pp. 2008--2021, 2018.

\bibitem{kumar2013towards}
Ajay Kumar and Cyril Kwong,
\newblock ``Towards contactless, low-cost and accurate 3d fingerprint identification,''
\newblock in {\em Proceedings of the IEEE Conference on Computer Vision and Pattern Recognition}, 2013, pp. 3438--3443.

\bibitem{cui2023monocular}
Zhe Cui, Jianjiang Feng, and Jie Zhou,
\newblock ``Monocular 3d fingerprint reconstruction and unwarping,''
\newblock {\em IEEE Transactions on Pattern Analysis and Machine Intelligence}, 2023.

\bibitem{kolberg2023colfispoof}
Jascha Kolberg, Jannis Priesnitz, Christian Rathgeb, and Christoph Busch,
\newblock ``Colfispoof: A new database for contactless fingerprint presentation attack detection research,''
\newblock in {\em Proceedings of the IEEE/CVF Winter Conference on Applications of Computer Vision}, 2023, pp. 653--661.

\bibitem{tolosana2019biometric}
Ruben Tolosana, Marta Gomez-Barrero, Christoph Busch, and Javier Ortega-Garcia,
\newblock ``Biometric presentation attack detection: Beyond the visible spectrum,''
\newblock {\em IEEE Transactions on Information Forensics and Security}, vol. 15, pp. 1261--1275, 2019.

\bibitem{hosseini2024intra}
Mohammad~Mehdi Hosseini, Mohammad~H Mahoor, Jason~W Haas, Joseph~R Ferrantelli, Anne-Lise Dupuis, Jason~O Jaeger, and Deed~E Harrison,
\newblock ``Intra-examiner reliability and validity of sagittal cervical spine mensuration methods using deep convolutional neural networks,''
\newblock {\em Journal of Clinical Medicine}, vol. 13, no. 9, pp. 2573, 2024.

\bibitem{fujio2018face}
Masakazu Fujio, Yosuke Kaga, Takao Murakami, Tetsushi Ohki, and Kenta Takahashi,
\newblock ``Face/fingerphoto spoof detection under noisy conditions by using deep convolutional neural network.,''
\newblock in {\em BIOSIGNALS}, 2018, pp. 54--62.

\bibitem{marasco2021fingerphoto}
Emanuela Marasco and Anudeep Vurity,
\newblock ``Fingerphoto presentation attack detection: Generalization in smartphones,''
\newblock in {\em 2021 IEEE International Conference on Big Data (Big Data)}. IEEE, 2021, pp. 4518--4523.

\bibitem{marasco2021deep}
Emanuela Marasco, Anudeep Vurity, and Asem Otham,
\newblock ``Deep color spaces for fingerphoto presentation attack detection in mobile devices,''
\newblock in {\em International Conference on Computer Vision and Image Processing}. Springer, 2021, pp. 351--362.

\bibitem{marasco2022late}
Emanuela Marasco and Anudeep Vurity,
\newblock ``Late deep fusion of color spaces to enhance finger photo presentation attack detection in smartphones,''
\newblock {\em Applied Sciences}, vol. 12, no. 22, pp. 11409, 2022.

\bibitem{purnapatra2023presentation}
Sandip Purnapatra, Conor Miller-Lynch, Stephen Miner, Yu~Liu, Keivan Bahmani, Soumyabrata Dey, and Stephanie Schuckers,
\newblock ``Presentation attack detection with advanced {CNN} models for noncontact-based fingerprint systems,''
\newblock pp. 1--6, 2023.

\bibitem{li2023deep}
Hailin Li and Raghavendra Ramachandra,
\newblock ``Deep features for contactless fingerprint presentation attack detection: Can they be generalized?,''
\newblock {\em arXiv preprint arXiv:2307.01845}, 2023.

\bibitem{purnapatra2023liveness}
Sandip Purnapatra, Humaira Rezaie, Bhavin Jawade, Yu~Liu, Yue Pan, Luke Brosell, Mst~Rumana Sumi, Lambert Igene, Alden Dimarco, Srirangaraj Setlur, et~al.,
\newblock ``Liveness detection competition--noncontact-based fingerprint algorithms and systems (livdet-2023 noncontact fingerprint),''
\newblock {\em arXiv preprint arXiv:2310.00659}, 2023.

\bibitem{taneja2016fingerphoto}
Archit Taneja, Aakriti Tayal, Aakarsh Malhorta, Anush Sankaran, Mayank Vatsa, and Rieha Singh,
\newblock ``Fingerphoto spoofing in mobile devices: a preliminary study,''
\newblock in {\em 2016 IEEE 8th International Conference on Biometrics Theory, Applications and Systems (BTAS)}. IEEE, 2016, pp. 1--7.

\bibitem{wasnik2018presentation}
Pankaj Wasnik, Raghavendra Ramachandra, Kiran Raja, and Christoph Busch,
\newblock ``Presentation attack detection for smartphone based fingerphoto recognition using second order local structures,''
\newblock in {\em 2018 14th International Conference on Signal-Image Technology \& Internet-Based Systems (SITIS)}. IEEE, 2018, pp. 241--246.

\bibitem{adami2023universal}
Banafsheh Adami, Sara Tehranipoor, Nasser Nasrabadi, and Nima Karimian,
\newblock ``A universal anti-spoofing approach for contactless fingerprint biometric systems,''
\newblock in {\em 2023 IEEE International Joint Conference on Biometrics (IJCB)}. IEEE, 2023, pp. 1--8.

\bibitem{adami2024contactless}
Banafsheh Adami, MohammadReza Hosseinzadehketilateh, and Nima Karimian,
\newblock ``Contactless fingerprint biometric anti-spoofing: An unsupervised deep learning approach,''
\newblock in {\em 2024 IEEE International Joint Conference on Biometrics (IJCB)}. IEEE, 2024, pp. 1--10.

\bibitem{liu2021swin}
Ze~Liu, Yutong Lin, Yue Cao, Han Hu, Yixuan Wei, Zheng Zhang, Stephen Lin, and Baining Guo,
\newblock ``Swin transformer: Hierarchical vision transformer using shifted windows,''
\newblock in {\em Proceedings of the IEEE/CVF international conference on computer vision}, 2021, pp. 10012--10022.

\bibitem{liu2022swin}
Ze~Liu, Han Hu, Yutong Lin, Zhuliang Yao, Zhenda Xie, Yixuan Wei, Jia Ning, Yue Cao, Zheng Zhang, Li~Dong, et~al.,
\newblock ``Swin transformer v2: Scaling up capacity and resolution,''
\newblock in {\em Proceedings of the IEEE/CVF conference on computer vision and pattern recognition}, 2022, pp. 12009--12019.

\bibitem{jia2016dynamic}
Xu~Jia, Bert De~Brabandere, Tinne Tuytelaars, and Luc~V Gool,
\newblock ``Dynamic filter networks,''
\newblock {\em Advances in neural information processing systems}, vol. 29, 2016.

\bibitem{dey2017gate}
Rahul Dey and Fathi~M Salem,
\newblock ``Gate-variants of gated recurrent unit (gru) neural networks,''
\newblock in {\em 2017 IEEE 60th international midwest symposium on circuits and systems (MWSCAS)}. IEEE, 2017, pp. 1597--1600.

\bibitem{woo2018cbam}
Sanghyun Woo, Jongchan Park, Joon-Young Lee, and In~So Kweon,
\newblock ``Cbam: Convolutional block attention module,''
\newblock in {\em Proceedings of the European conference on computer vision (ECCV)}, 2018, pp. 3--19.

\bibitem{kingma2014adam}
Diederik~P Kingma and Jimmy Ba,
\newblock ``Adam: A method for stochastic optimization,''
\newblock {\em arXiv preprint arXiv:1412.6980}, 2014.

\bibitem{lin2017focal}
Tsung-Yi Lin, Priya Goyal, Ross Girshick, Kaiming He, and Piotr Doll{'a}r,
\newblock ``Focal loss for dense object detection,''
\newblock in {\em Proceedings of the IEEE international conference on computer vision}, 2017, pp. 2980--2988.

\bibitem{hadsell2006dimensionality}
Raia Hadsell, Sumit Chopra, and Yann LeCun,
\newblock ``Dimensionality reduction by learning an invariant mapping,''
\newblock in {\em 2006 IEEE Computer Society Conference on Computer Vision and Pattern Recognition (CVPR'06)}. IEEE, 2006, vol.~2, pp. 1735--1742.

\bibitem{sankaran2015smartphone}
Anush Sankaran, Aakarsh Malhotra, Apoorva Mittal, Mayank Vatsa, and Richa Singh,
\newblock ``On smartphone camera based fingerphoto authentication,''
\newblock in {\em 2015 IEEE 7th International Conference on Biometrics Theory, Applications and Systems (BTAS)}. IEEE, 2015, pp. 1--7.

\end{thebibliography}
}

\end{document}